\documentclass[journal]{IEEEtran}
\usepackage{cite}
\usepackage{amsmath,amssymb,amsfonts}
\usepackage{algorithmic}
\usepackage{graphicx}
\usepackage{textcomp}
\usepackage{xcolor}
\usepackage[numbers,sort&compress]{natbib}
\usepackage{hyperref}
\usepackage{soul}

\ifCLASSINFOpdf
\else
\fi
\begin{document}
%
\title{Automatic Depression Detection via Learning and Fusing Features from Visual Cues}
%
%
%

\author{Yanrong~Guo,
        Chenyang~Zhu,
        Shijie~Hao,
        and~Richang~Hong
\thanks{Y. Guo, C. Zhu, S. Hao, and R. Hong are all with Key Laboratory of Knowledge Engineering with Big Data (Hefei University of technology), Ministry of Education and School of Computer Science and Information Engineering, Hefei University of Technology (HFUT), 230009 China e-mail: yrguo@hfut.edu.cn.}
\thanks{This work has been submitted to the IEEE for possible publication. Copyright may be transferred without notice, after which this version may no longer be accessible.}
}
\maketitle

\begin{abstract}
Depression is one of the most prevalent mental disorders, which seriously affects one's life. Traditional depression diagnostics commonly depends on rating with scales, which can be labor-intensive and subjective. In this context, Automatic Depression Detection (ADD) has been attracting more attention for its low cost and objectivity. ADD systems are able to detect depression automatically from some medical records, like video sequences. However, it remains challenging to effectively extract depression-specific information from long sequences, thereby hindering a satisfying accuracy. In this paper, we propose a novel ADD method via learning and fusing features from visual cues. Specifically, we firstly construct Temporal Dilated Convolutional Network (TDCN), in which multiple Dilated Convolution Blocks (DCB) are designed and stacked, to learn the long-range temporal information from sequences. Then, the Feature-Wise Attention (FWA) module is adopted to fuse different features extracted from TDCNs. The module learns to assign weights for the feature channels, aiming to better incorporate different kinds of visual features and further enhance the detection accuracy. Our method achieves the state-of-the-art performance on the DAIC\_WOZ dataset compared to other visual-feature-based methods, showing its effectiveness.
\end{abstract}


%
\IEEEpeerreviewmaketitle

\section{Introduction}

\IEEEPARstart{D}{epression} is a major mental disorder in nowadays society. It has a negative impact on one's life, even leading to suicide in extreme conditions \cite{2019Providing}. What's more urgent, despite that depression has become one of the most prevalent psychiatric disorders, the number of people suffering from this disease is still on the rise. Luckily, the symptoms of depression can be be alleviated on the premise of in-time diagnosis. Therefore, accurate depression diagnosis has become the key factor. Traditional depression diagnostics mostly rely on subjective rating with scales, such as the Eight-item Patient Health Questionnaire depression scale (PHQ-8) \cite{kroenke2009the_14}, where patients with scores no smaller than 10 are considered as depressed. It needs the participation of experienced psychologists, which can be labor-intensive. However, subjective bias may still exist, therefore leading to misdiagnosis. Besides, different from common physical illness, the severity of mental disorders may vary at different time \cite{kazdin2011rebooting,mitchell2009clinical}, which brings extra challenges to the diagnostics. Consequently, automatic depression detection (ADD) has been drawing increasing attention. It is typically realized through a pattern classification paradigm based on one's verbal or facial behavior data. Obviously, this process is more objective, while requiring much fewer human interventions.

	\begin{figure*}[htpb]
		\centering
		\includegraphics[width=\textwidth]{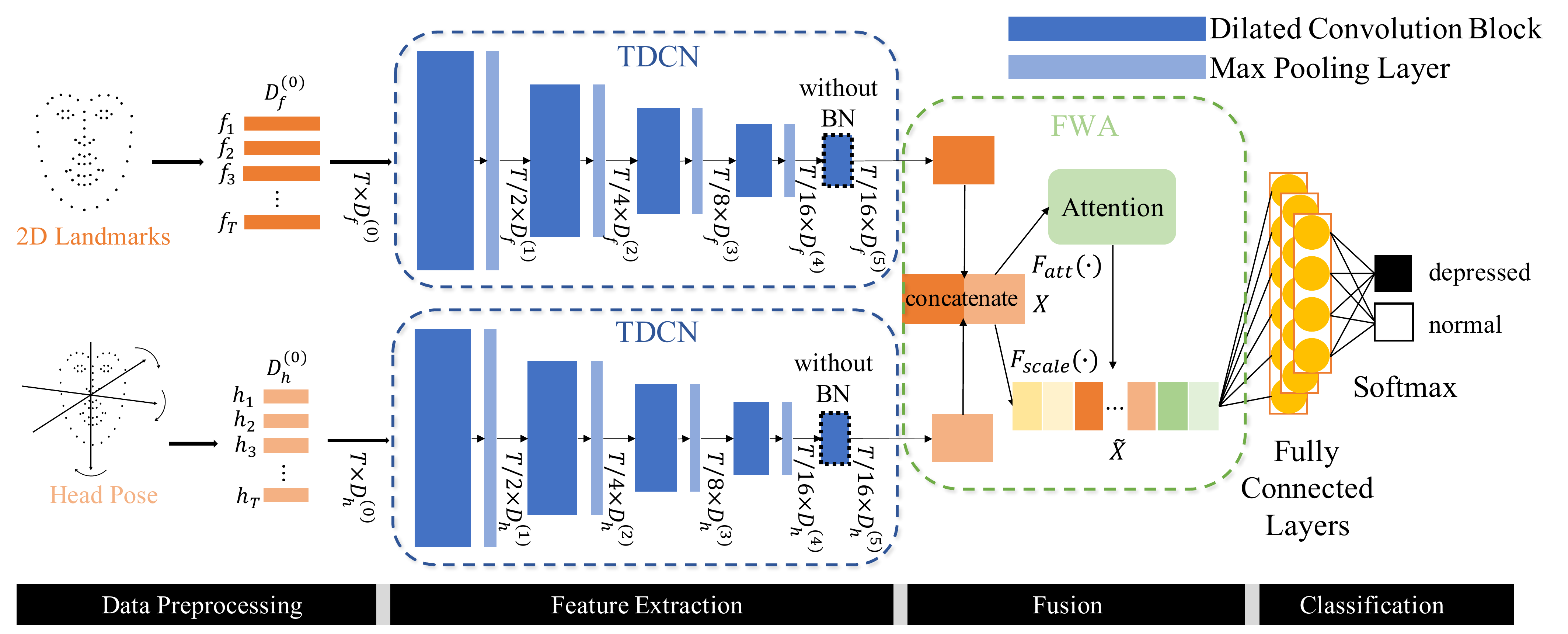}
		\caption{The framework of our model. The sequences of 2D facial landmarks and head-pose are taken as the inputs. We design Temporal Dilated Convolutional Network (TDCN), which consists of Dilated Convolutional Blocks (DCB) and max pooling layers, to extract the depression-specific features. With the features obtained from the two TDCN branches, Feature-Wise Attention (FWA) is then adopted, which assigns higher weights to more valuable parts of the concatenated features. As last, the fused features are fed into a simple classifier, which is composed of three fully connected layers and a Softmax layer. The outputs are the probabilities of being depressed and non-depressed.}
		\label{framework}
	\end{figure*}

	Images or videos provide rich visual cues of mental conditions. Compared to common people, patients suffering from depression usually have different facial actions, such as glazed expressions, abnormal head movements \cite{2000Motor,1993Psychomotor}. Due to the importance of these facial emotions or dynamic changes, visual data become a main information source for detecting depression. However, depression does not appear as instant behaviors. Instead, it is not distinguishable until a relatively long observation time is given. Thus, the detection model usually requires a large perceptive field to explore the depression-specific information. Therefore, as for depression detection based on visual cues, it is necessary to take care of the dynamic characteristics of data, which can be called temporal information. Some works try to solve this issue with LSTM \cite{isbi_10} or TCN \cite{ccnn_9} and achieve promising results. Nevertheless, limitations still exist in two aspects. On the one hand, they are less suitable for processing the overlong sequences. On the other hand, they prefer to using a single visual cue, ignoring the complementarity between different kinds of visual information.

	To tackle these limitations, we propose a novel ADD method based on visual cues in this paper. The whole framework of our method can be seen in \autoref{framework}, which has two main modules: Temporal Dilated Convolution Network (TDCN) and Feature-Wise Attention (FWA). Specifically, TDCN extracts depression-specific information by fulling utilizing the dilated convolutions of various receptive fields. FWA further enhances the feature representation ability by assigning different weights to the feature channels. We achieve the state-of-art performance compared to other visual-based works, e.g. 0.800 F1-score, on the validation set of the DAIC\_WOZ dataset. Generally, the contributions of our work are two-fold:

	\begin{itemize}
		\item We propose Temporal Dilated Convolutional Network (TDCN) to effectively extract the temporal information from long video sequences. Within TDCN, two parallel dilated convolutional modules are applied to learn useful temporal information for detecting depression, and the max pooling layers are adopted to solve the problem of overlength.
	
		\item We construct the Feature-Wise Attention (FWA) module to fuse the learned features from different TDCN branches. With the attention module, our method is able to further highlight the important part of the learned features, thereby enhancing the ADD accuracy.
	\end{itemize}

	\section{Related Work}
	
	Many methods have been developed for the task of automatic depression detection. Some methods are based on a single modality like audio \cite{muzammel2020audvowelconsnet,saidi2020hybrid,audio2021}, text \cite{text1,text2} or visual cues \cite{du2019encoding_8,isbi_10}, while others combine at least two modalities that tend to achieve a higher accuracy \cite{2021ta2,2021ta1}.

	 Early ADD works typical follow the traditional roadmap of a pattern recognition task, i.e., hand-crafted features are firstly designed based on visual data, and then novel classifiers are constructed to achieve the depression detection. Either of these two steps can be the key factors in the ADD research. For example, Wang et al. \cite{wang2008automated_1} design probabilistic classifiers for the purpose of judging the mental state of candidates through variations of their facial expressions. Yang et al. \cite{yang2016decision_2} propose a gender-specific decision tree for depression classification. As for the visual cues, geometric features from 2D facial landmarks are extracted. Besides, gaze and pose features and emotion evidence measures provided by AVEC2016 \cite{2016avec_3} are also added. Despite the obtained promising results, they take the average of each feature as input without considering the long-term correlations, inevitably losing the temporal information. Some other methods also have the similar issue. Pampouchidou et al. \cite{pampouchidou2016depression_4} extract low-level features such as Landmark Motion History Images (LMHI) and Landmark Motion Magnitude (LMM) from 2D facial landmarks, both of which encode the motion of features. However, the temporal information of other visual cues are mostly neglected. In a word, the works based on the traditional pattern recognition paradigm have a common problem, that is, the features extraction process are usually constructed based on some prior knowledge or ad hoc rules, which can be less comprehensive and robust for the ADD task.

	 Recently, more and more ADD methods based on the deep learning models have been proposed, as they are advantageous for solving the issue of hand-crafted features. To learn the correlation and dynamic variation between time frames, the Long-Short Term Memory (LSTM) neural networks \cite{lstm} or other Recurrent Neural Networks (RNN) for such temporal data can be usually adopted. For example, both \cite{hanai2018detecting_6} and \cite{ray2019multi_7} choose LSTM as their backbone networks. However, the performance of LSTM degrades when sequences become too long \cite{lea2016temporal_12}. In other words, LSTM are not good at handling overlength sequences in practice. Besides, the number of parameters becomes huge when they receive overlong sequences, possibly leading to speed reduction and overfitting in the training process. One feasible solution is to shorten the input length directly by sampling. Wang et al. \cite{isbi_10} introduce Sampling Slicing Long Short Term Memory Multiple Instance Learning (SS-LSTM-MIL), a method based on multiple instance learning with 2D facial landmarks as input. Its sampling strategy is that only frames when patients are speaking are chosen. It outperforms other visual-based methods in the ADD task. Another solution is to do embedding first to unify the length of sequences or feature dimensions. Du et al. \cite{du2019encoding_8} propose DepArt-Net for depression recognition, which encodes four visual cues into long-term representations and then applies atrous residual temporal convolution with attentive temporal pooling. Haque et al. \cite{ccnn_9} leverage a casual convolutional neural network, which is a TCN \cite{tcnn} actually, with a sentence-level summary embedding to detect the major depressive disorder. However, TCN still has some disadvantages that are analyzed in the next section.

	Different from the above methods, our work is designed to better relieve the overlength issue. First, in order to decrease the computational complexity of our model, we split samples into several sequences with fixed sizes. Second, we propose the TDCN model which fully utilizes the information from perceptive fields of various sizes.
	
	\section{Method}
	\subsection{Overview}
	Generally, as shown in \autoref{framework}, the overall framework of our method includes four parts: data preprocessing, feature extraction, fusion and classification. Specifically, we firstly divide samples into sequences with fixed shape. Then, we use two TDCN branches to learn discriminate features for detecting depression. In the third part, the learned features are fused by FWA, which assigns weights to different feature channels. Finally, as the main part of the whole model, the details of TDCN and FWA are described in the following.

	\subsection{Temporal Dilated Convolution Network}

	\autoref{framework} shows the general structure of the proposed TDCN. Generally, TDCN is a multi-layer structure, consisting of five Dilated Convolutional Blocks (DCB) and four max pooling layers. On the one hand, within a certain TDCN layer, DCB explores the useful information at different perceptive ranges. On the other hand, along the TDCN pipeline, the max pooling layers keep on shrinking the feature resolutions and gradually extract the most important responses. Therefore, TDCN can be seen as a feature learning module that extracts depression-specific information from multiple scales.
	
\begin{figure}[htpb]
		\centering
		\includegraphics[width=\linewidth]{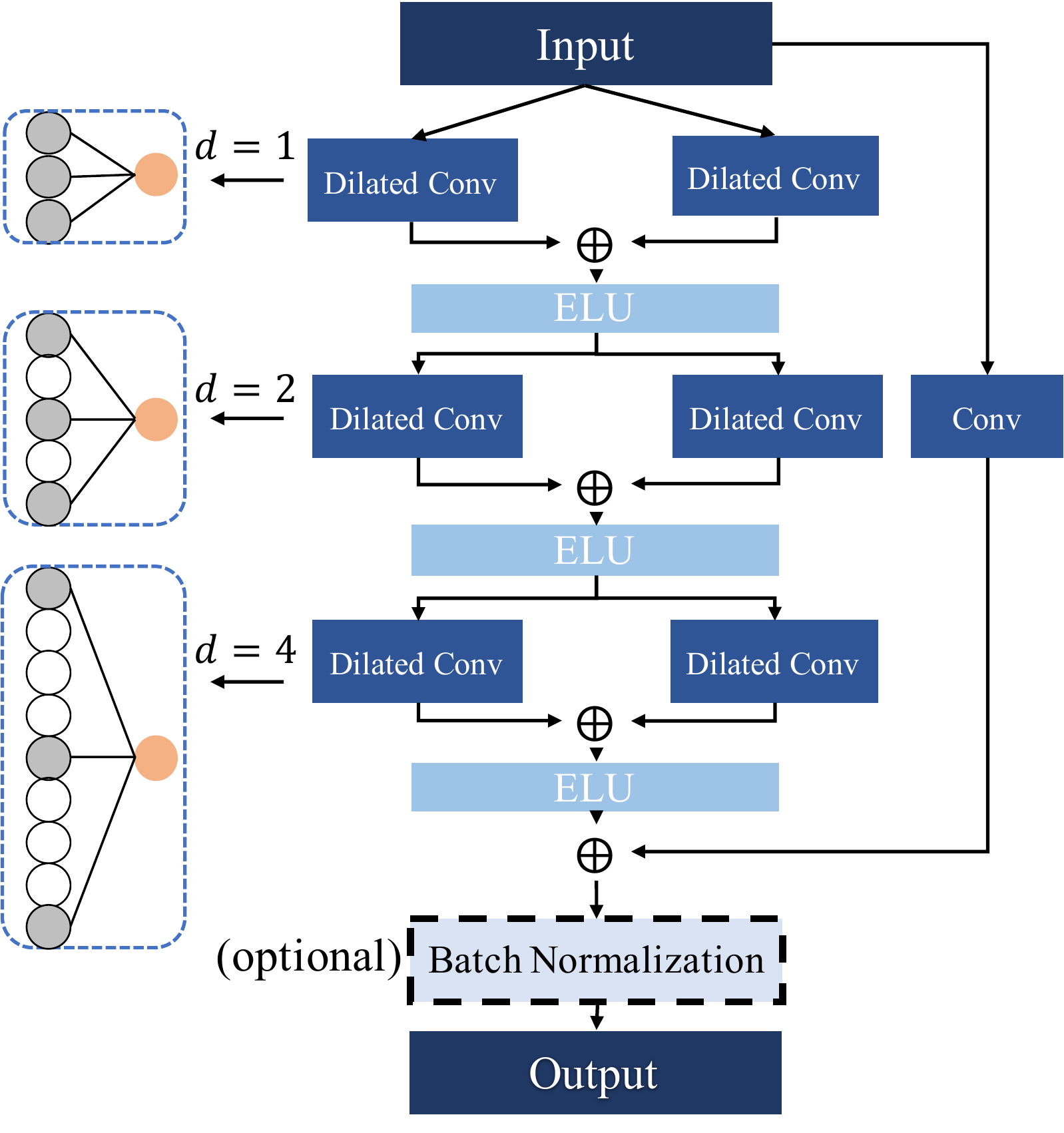}
		\caption{the structure of a Dilated Convolutional Block}
		\label{dcb}
	\end{figure}

	As shown in \autoref{framework}, DCB is the main components of a TDCN module, which is described in the following. In \autoref{dcb}, we can see that two parallel dilated convolution paths are formulated, which extract features with different dilation factors. Given an input $X=[x_1;x_2;\cdots ;x_T]\in \mathbb{R}^{T \times D}$ where $T$ is the time steps and $D$ is the feature dimension, the operation of dilated convolution can be expressed as:
	
	\begin{equation}
	    F(t)=\sum_{i=0}^{k-1} filter(i)\cdot x_{t+d\cdot (i-1)}+b
	\end{equation}
	where $d$ is the dilation factor, $k$ is the kernel size and $b$ is the bias. Zero padding is adopted to keep the same shape of its input and output. As for our ADD application, the dilation factor doubles by a factor of 2, aiming to obtain temporal information from different time spans along the path. At different dialtion factors, the two paths join together by means of the summation and activation operations. We choose ELU \cite{elu} as our activation function, which is defined as:
	\begin{equation}
	    f_{ELU}(x)=
	    \begin{cases}
	        x&\text{if $x\ge0$}\\
	        e^x-1&\text{if $x<0$}
	    \end{cases}
	\end{equation}
	
	To avoid degradation issue when the network goes deeper, a residual block \cite{resnet} is added in our DCB. In addition, to keep the same shape of the element-wise addition in shortcut connections, we add a 1D convolutional layer with a kernel size of 1 in every DCB. At the end of DCB, batch normalization layer is applied to accelerate the training process. Besides, it is able to alleviate the problem of gradient vanishing. Of note, we remove it from the last DCB in TDCN to reserve the distribution of different features.

	We add a max pooling layer after each DCB, except the last one in TDCN. With this operation, its output tensor obtains a broader receptive field. Its aim is to gradually aggregate the most important information from the long sequence. In addition, Max pooling layers also cut down the length of sequences and retain the important part, reducing the complexity of our model.

\begin{figure}[htpb]
		\begin{minipage}{1\linewidth}
			\vspace{3pt}
			\centerline{
				\includegraphics[width=\textwidth]{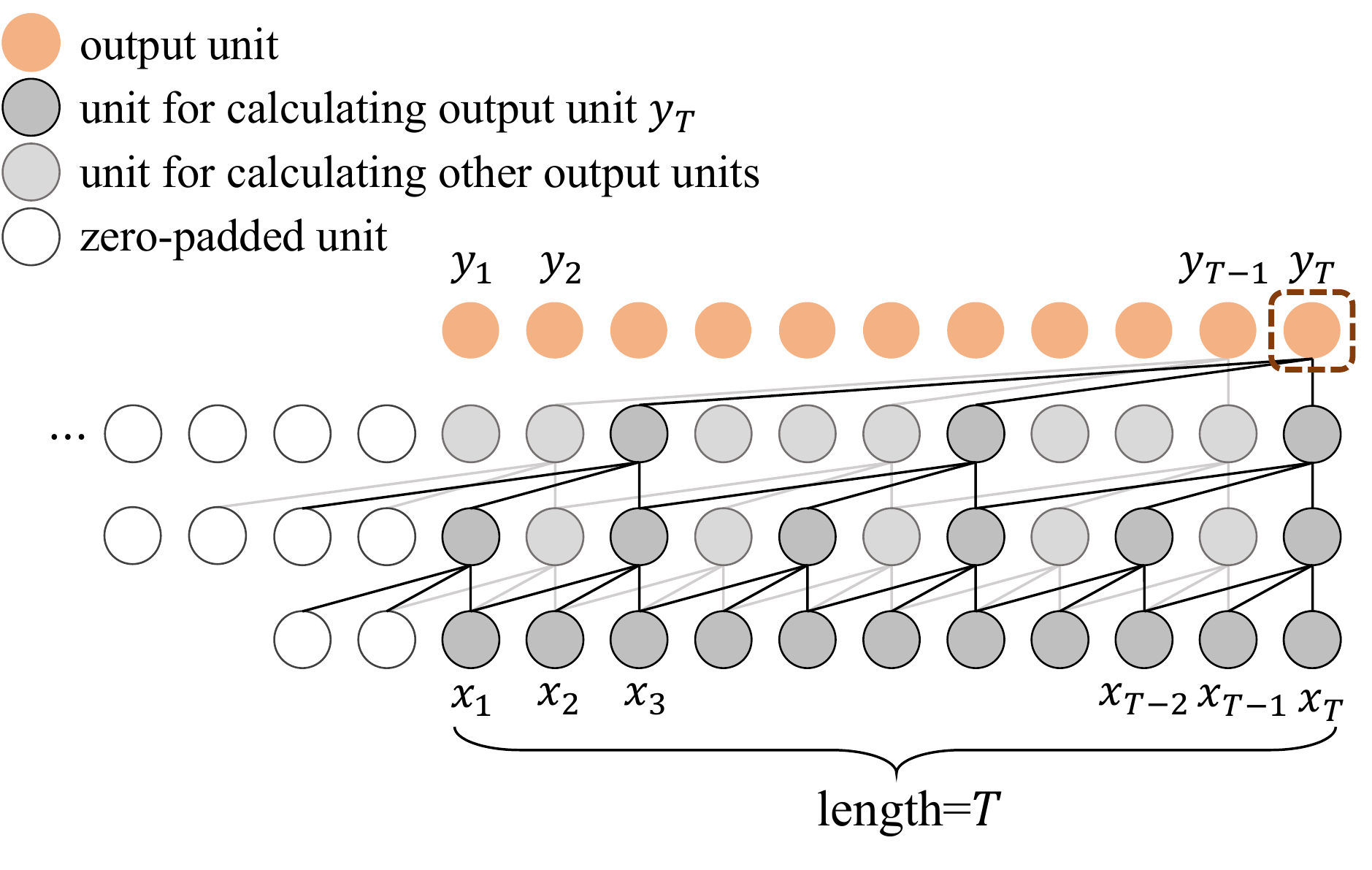}}
			\centerline{(a) the basic structure of TCN}
			
			\vspace{3pt}
			\centerline{
				\includegraphics[width=\textwidth]{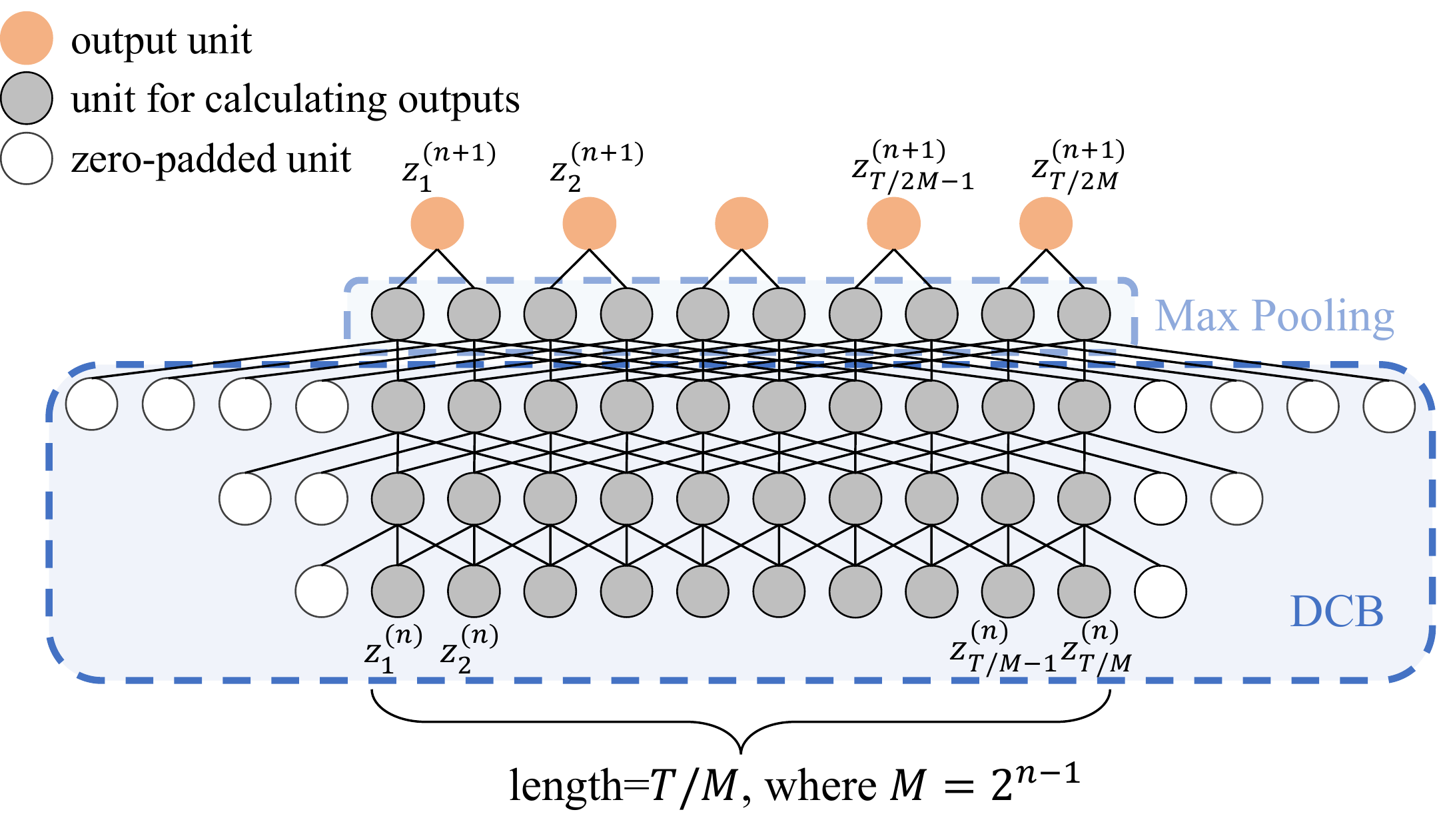}}
			\centerline{(b) $n$-th DCB+max pooling layer in TDCN}
		\end{minipage}
		
		\caption{The structure of TCN and TDCN}
		\label{comparison}
	\end{figure}

	In the following, we further analyse the difference between Temporal Convolutional Networks (TCN) \cite{tcn1} and our TDCN module. To enlarge the receptive field of the traditional convolution, Bai et al. \cite{tcnn} develop a new architecture for TCN by employing dilated convolutions. However, as sequence-to-sequence architectures, TCN is less suitable for handling overlong video sequences due to the insufficient receptive field. Take \autoref{comparison}(a) for example, given an input $X=[x_1;x_2; \cdots; x_T]\in \mathbb{R}^{T\times D_{in}}$, the length of TCN's output $Y=[y_1;y_2;\cdots;y_T]\in \mathbb{R}^{T\times D_{out}}$ is same as $X$. Therefore, the size of $Y$ can be large in applications in which the input is also very long. In this context, $y_T$ in $Y$ can be selected as the input of the final classifier. The rationale comes from the use of causal convolutions, such as the connection between units of different layers in \autoref{comparison}(a). Based on this construction, the network is expected to be deeper when a larger receptive field is needed in our ADD application. This potentially brings in the side effects such as oversize dilation factor and more computational cost, which may lower the detection accuracy of the TCN-based model, as well as the efficiency.

   Our Temporal Dilated Convolutional Network (TDCN) relieves the above limitation by using a different structure. \autoref{comparison}(b) shows the $n$-th DCB+max pooling layer in our TDCN. In the DCB part, we incorporate multiple dilated convolutions with increasing dilated factors ($d=1, 2, 4$). Then a max pooling layer is followed, shrinking the feature length and naturally enlarging the receptive field for its neighboring ($n$+1)-th DCB, of which the input size has been reduced into $T/2M$ ($M=2^{n-1}, n=1\cdots5$). Compared with the TCN structure, TDCN better controls the receptive field of long sequences in terms of flexibility and efficiency. This is also empirically validated by the experimental results.

	\subsection{Feature-Wise Attention}
\begin{figure}[htpb]
		\centering
		\includegraphics[width=\linewidth]{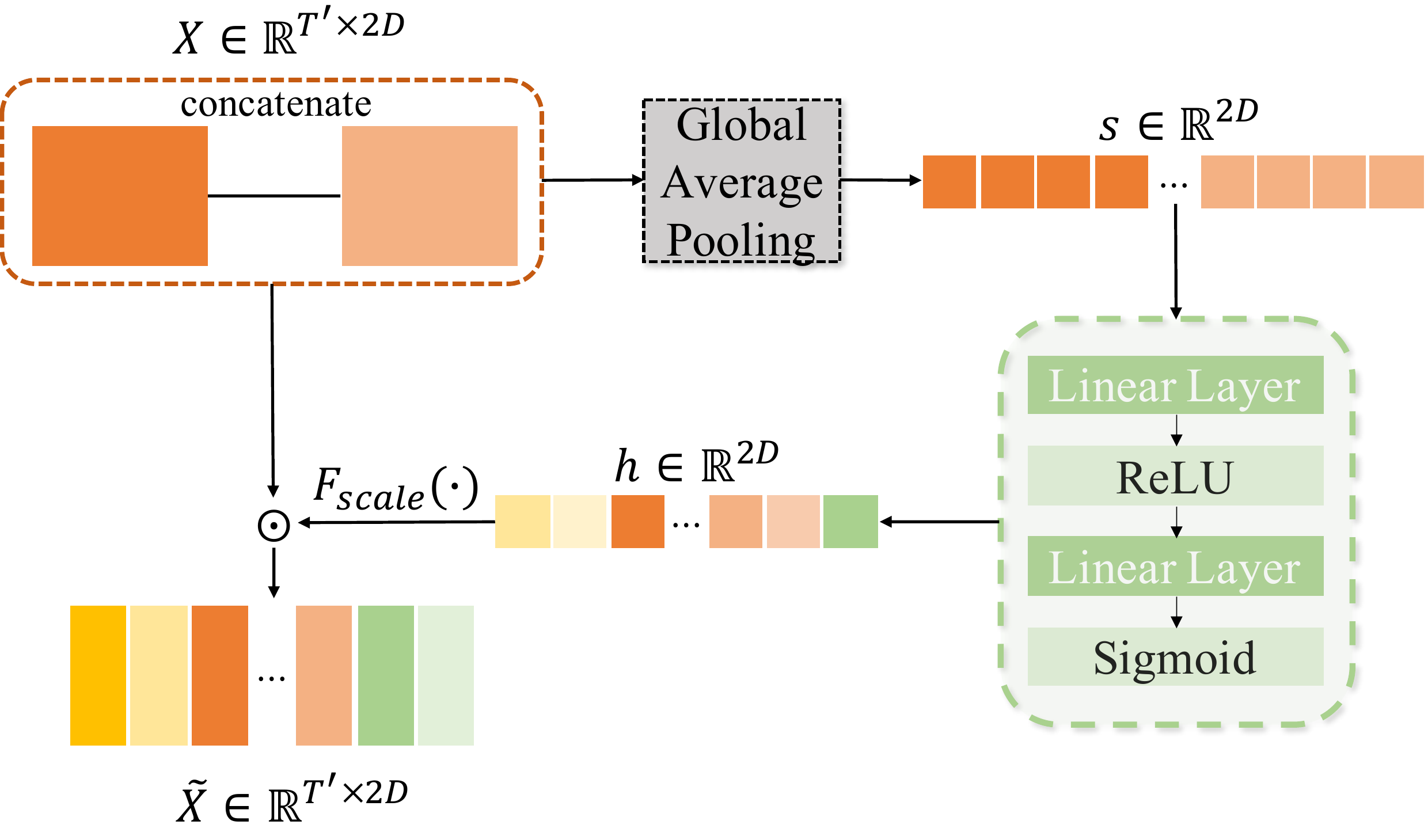}
		\caption{The structure of Feature-Wise Attention Module}
		\label{fwa}
	\end{figure}

	In the visual-based ADD task, various visual cues represent the interviewee's status from different views. For example, the sequence of landmark positions represents one's facial morphing, while the pose vector sequence represents how one's face orientates during the interviewing process. They all potentially contribution to the final inferring process in our ADD research. To make the most of visual information, we construct the FWA module, inspired by \cite{hu2018squeeze_13}, to effectively combine the different features together. We first concatenate the features learned from different TDCN branches directly, obtaining $X\in\mathbb{R}^{T\times k D}$, where $D$ is the feature dimensions and $k$ is the number of TDCN branches. They are then fed into the feature-wise attention module to learn the weight of each feature dimension based on its significance. The details of the FWA module can be seen in \autoref{fwa}. The global average pooling is applied to achieve a feature-wise vector $s\in\mathbb{R}^{kD}$, obtained as:
	\begin{equation}
		s_j=\frac{1}{T}\sum^{T-1}_{i=0}x_{i,j}
	\end{equation}
	where $x_{i,j}$ is the unit of $i$-th time-step and $j$-th feature dimension of $X$. After that, two linear layers, as well as a ReLU operation, are imposed on $s$. Subsequently, a sigmoid activation is applied to learn the non-linear correlation between the features. Formally, this process can be expressed as:
	\begin{equation}
		h=\mathcal{F}_{att}(s,W)=\sigma_{sigmoid}(W_2(f_{ReLU}(W_1 s)))
	\end{equation}
	where $W_1$ and $W_2$ are the weighted matrices of linear layers. $h\in\mathbb{R}^{kD}$ reveals the importance of feature channels. The output of feature-wise attention $\tilde{X}$ is obtained by broadcasting $h$ to the same shape of $X$, and then element-wise product $X$:
	\begin{equation}
		\tilde{X}=\mathcal{F}_{scale}(x,h)=X\odot\tilde{H}
	\end{equation}
	where $\tilde{H}\in\mathbb{R}^{T\times kD}$ is the rescaled output of $h$. The number of visual cues $k$ is set to 2 in our work.
	
	\section{Experiments}

	\subsection{Dataset and Implementation details}
	In our research, we use the Distress Analysis Interview Corpus Wizard-of-Oz dataset (DAIC\_WOZ) \cite{daic} to evaluate the effectiveness of our model. The Distress Analysis Interview Corpus (DAIC) collects interviews of semi-structured clinical. The Wizard-of-Oz interviews use a virtual avatar named Ellie to interact with patients, asking fundamental questions, and collecting audio, video, and depth sensor recordings in the meanwhile. In general, this dataset includes audio and video recordings and text transcripts from audio data. The number of samples for training/validation/testing is 107/35/47 respectively. As our model needs inputs with the same shape, it is necessary to re-sample the sequence and set them into a unified length. In the meanwhile, the overall length of one input is expected to be relatively long so as to preserve the temporal information. To this end, we empirically perform the head-first re-sample to all the sequences in the dataset, and set the length of every sample as 5000.

	\begin{table*}[htbp]
		\caption{Comparisons on DAIC\_WOZ validation set}
		\label{validation}
		\centering
		\begin{tabular}{cccccc}
			\hline
			\multicolumn{1}{c|}{Method}            & \multicolumn{1}{c|}{Feature} & Accuracy & Recall & Precision & F1-score \\ \hline
			\multicolumn{1}{c|}{SVM \cite{2016avec_3}}               & \multicolumn{1}{c|}{V}        & -        & 0.428  & 0.600     & 0.500    \\
			\multicolumn{1}{c|}{GSM \cite{williamson2016detecting}} & \multicolumn{1}{c|}{AUs}        & -        & -      & -         & 0.530    \\
			\multicolumn{1}{c|}{CNN \cite{cnn}}      & \multicolumn{1}{c|}{AUs+Gaze+Pose}        & -        & 0.583   & 0.636      & 0.609     \\
			\multicolumn{1}{c|}{SGD-SVM \cite{nasir}}      & \multicolumn{1}{c|}{3D Landmarks+Gaze + Pose}        & -        & 0.71   & 0.56      & 0.63     \\
			\multicolumn{1}{c|}{Topic \cite{gong2017topic_5}}       & \multicolumn{1}{c|}{A+L+AUs}      & -        & -      & -         & 0.70     \\
			\multicolumn{1}{c|}{C-CNN \cite{ccnn_9}}             & \multicolumn{1}{c|}{A+L+3D Landmarks}      & -        & \textbf{0.833}  & 0.714     & 0.769    \\
			\multicolumn{1}{c|}{SS-LSTM-MIL \cite{isbi_10}}       & \multicolumn{1}{c|}{2D Landmarks}        & -        & 0.75   & \textbf{0.818}     & 0.783    \\ \hline
			\multicolumn{1}{c|}{Ours}              & \multicolumn{1}{c|}{2D Landmarks + Pose}        & \textbf{0.857}    & \textbf{0.833}  & 0.770     & \textbf{0.800}    \\ \hline
		\end{tabular}
	\end{table*}

	To protect the privacy of participants, DAIC\_WOZ does not release the raw video recordings. Instead, several visual cues extracted from video recordings with OpenFace toolkit \cite{baltrusaitis2016openface_11} are provided, including 68 2D/3D facial landmarks, Action Units (AUs), gaze vector, head-pose vector and Histogram of Oriented Gradients (HOG) features. 2D facial landmarks record the two-dimensional coordinates of 68 facial feature points with a total of 136 dimensions. Pose features contain 6 dimensions, including X, Y, Z, Rx, Ry and Rz, where X, Y, Z are the position coordinates of the head, and Rx, Ry, Rz are the head rotation coordinates. Gaze vectors record the eye-gaze directions. AUs record the existence of 20 action units. HOG is a feature descriptor used for object detection, of which the size is too huge for our application. As for the rest features, we select two of them to validate the effectiveness of the FWA module, i.e. 2D facial landmarks and head-pose features. Of note, we empirically find that incorporating three or more kinds of features does not gain extra detection accuracy, which would be discussed in the analysis of our method.
	
	The implementation details are described in the following. Depressed subjects are labeled as 1 (being positive), and non-depressed subjects are labeled as 0 (being negative). The feature dimensions of Dilated Convolutional Blocks are ${256,256,128,64,64}$ for 2D facial landmark features, and ${128,64,256,128,64}$ for the head-pose features. During the training stage, we choose SGD as the optimizer whose learning rate is 2e-5 and momentum rate is 0.9. The mini-batch size is set as 8.

	\subsection{Evaluation Metrics}

	As shown in Eq. (\ref{metrics}), we choose the metrics commonly used in classification tasks, i.e. accuracy, precision, recall and F1-score, to evaluate performance. TP, TN, FP and FN denote True Positive, True Negative, False Positive and False Negative, respectively. As for our application, for example, TP means the number of the depressed subjects that are predicted as depressed, and TN means the number of the non-depressed subjects that are predicted as non-depressed.
	
			\begin{equation}
	    \label{metrics}
		\begin{split}
			Accuracy= &\frac{TP+TN}{TP+TN+FP+FN}\\
			Recall= &\frac{TP}{TP+FN}\\
			Precision= &\frac{TP}{TP+FP}\\
			F1-score= &2\times\frac{Precision\times Recall}{Precision+Recall}
		\end{split}
	\end{equation}
	
	Based on these, Accuracy represents the percentage of correctly classified samples in the total number of samples. Recall represents the percentage of the positive samples that are correctly predicted. Precision represents the percentage of correctly predicted positive subjects in the subjects predicted as positive. F1-score jointly considers the metrics of Precision and Recall. In particular, Recall is clinically important, as it reflects a model's capability of finding out patients suffering from depression from all participants. In another word, the price that the detection model misses one depressed subject is much higher. In addition, F1-score also matters in our application as this metric jointly considers Recall and Precision.

	\subsection{Comparison with Other Methods}
	We compare our method with the related ADD methods utilizing the visual modality. Considering that most works evaluate their methods on the validation set, we split a part of the original training set for tuning the hyperparameters. Then we report our results on the validation set for a fair comparison. The results are shown in \autoref{validation}, in which 'A', 'V', 'L' denotes audio, visual and textual modality, respectively.
	
    From the table, we can see several general trends. First, the overall results demonstrate the usefulness of the visual cues in the ADD task, both for the multi-modality and single-modality methods. Second, two factors contribute to the success of a ADD model, that is, the subtle construction of learning models and the effective incorporation of features. As for the comparison between the related methods and ours, we have the following detailed observations. First, we achieve very competitive performance among all the methods. Our method achieves the best performance in the metrics of Accuracy, Recall and F1-score, and achieves the second highest score on the Precision metric. Second, the results of our method surpass the ones based on multi-modalities, demonstrating that our method is able to fully explore the useful visual information for accurate depression detection.

	\subsection{Analysis of Our Method}
	
	We conduct more experiments to investigate the influence and the effectiveness of our method's elements. Of note, in these experiments, we follow the original partition of the dataset (107/35/47), and tune hyperparameters with which our model performs best on the original validation set. Consequently, the results on the validation set are different from those in \autoref{validation}.
	
	\begin{table}[htbp]
		\caption{Results of The TDCN With Single Feature Compared to TDCN with FWA}
		\label{test}
		\centering
		\begin{tabular}{ccccc}
			\hline
			\multicolumn{5}{c}{Validation Set}                                                                                     \\ \hline
			\multicolumn{1}{c|}{Feature}                     & Accuracy        & Recall         & Precision      & F1-score       \\ \hline
			\multicolumn{1}{c|}{AUs}                          & 0.800           & 0.500          & \textbf{0.857} & 0.632          \\
			\multicolumn{1}{c|}{Gaze}                         & 0.800           & 0.500          & \textbf{0.857} & 0.632          \\
			\multicolumn{1}{c|}{Pose}                         &0.800           & 0.667          & 0.727          & 0.696          \\
			\multicolumn{1}{c|}{2D Landmarks}                 & 0.800  & 0.750 & 0.692          & 0.720 \\
			\multicolumn{1}{c|}{\textbf{2D Landmarks+Pose}} & \textbf{0.857}           & \textbf{0.917} & 0.733          & \textbf{0.815}          \\ \hline
			\multicolumn{5}{c}{Test Set}                                                                                        \\ \hline
			\multicolumn{1}{c|}{Feature}                     & Accuracy        & Recall         & Precision      & F1-score       \\ \hline
			\multicolumn{1}{c|}{AUs}                          & 0.638           & 0.357          & 0.385          & 0.370          \\
			\multicolumn{1}{c|}{Gaze}                         & 0.596           & 0.214          & 0.273          & 0.240          \\
			\multicolumn{1}{c|}{Pose}                         & \textbf{0.660}           & 0.214          & 0.375          & 0.273          \\
			\multicolumn{1}{c|}{2D Landmarks}                 & 0.596 & 0.214          & 0.273 & 0.240          \\
			\multicolumn{1}{c|}{\textbf{2D Landmarks+Pose}} & \textbf{0.660}           & \textbf{0.643} & \textbf{0.450}          & \textbf{0.530} \\ \hline
		\end{tabular}
	\end{table}
	
	We first show the necessity of incorporating different kinds of visual cues. As is shown in \autoref{test}, we compare the result of our model with its intermediate version, which only uses one single feature with one TDCN branch. We can observe a general trend that the performance of incorporating two kinds of features is better than the performance based on a single kind of feature. Specifically, the Recall metric is largely improved in both the validation set and the test set. In another word, our model is able to detect more depressed individuals from all the subjects, which can be meaningful in clinical applications. In contrast, compared with the performance of Accuracy and Precision, the Recall scores of the single-feature models are relatively low, especially for the results of the test set, which makes them almost useless in clinical applications. These results again empirically validate the effectiveness of incorporating different kinds of visual cues in the ADD task.

    \begin{table}[htbp]
		\caption{Results of Our Model With Different fused features on DAIC\_WOZ validation set}
		\label{fused features}
		\centering
		\begin{tabular}{ccccc}
			\hline
			\multicolumn{1}{c|}{\scriptsize Fused Features}                & Accuracy       & Recall         & Precision      & F1-score       \\ \hline
			\multicolumn{1}{c|}{\scriptsize AUs+Gaze}        &0.800           &0.417          &\textbf{1.00}       &0.588           \\
			\multicolumn{1}{c|}{\scriptsize Pose+AUs}      & 0.800          & 0.750           & 0.692 & 0.720          \\
			\multicolumn{1}{c|}{\scriptsize Pose+Gaze}      & 0.800          & 0.500          & 0.857 & 0.632          \\
			\multicolumn{1}{c|}{\scriptsize Landmarks+AUs}      & 0.829          & \textbf{0.917}          & 0.688 & 0.786          \\
			\multicolumn{1}{c|}{\scriptsize Landmarks+Gaze}      & 0.771          & \textbf{0.917}          & 0.611 & 0.733          \\
			\multicolumn{1}{c|}{\scriptsize \textbf{Landmarks+Pose}}      & \textbf{0.857}          & \textbf{0.917}          & 0.733 &\textbf{0.815}       \\
            \multicolumn{1}{c|}{\scriptsize Landmarks+Pose+Gaze}      & 0.743          & 0.583          & 0.634 & 0.609 \\
            \multicolumn{1}{c|}{\scriptsize Landmarks+AUs+Pose}      & 0.686          & 0.583          & 0.539 & 0.560 \\
            \multicolumn{1}{c|}{\scriptsize Landmarks+AUs+Pose+Gaze}      & 0.686          & 0.333          & 0.571 &0.421 \\\hline
		\end{tabular}
	\end{table}

	As a step further, as shown in \autoref{fused features}, we report the results of the different feature combinations. In general, compared with the results in \autoref{test}, we can see that almost all the combinations obtain better performance than the single-feature configurations. Specifically, we observe that incorporating the landmark feature is important to the final results, especially for the metrics of Recall and F1-score. The reason is that the facial landmarks provide more details of how one's facial organs morph in a fine scale. Based on that, we empirically find that the 2D Landmarks+Pose configuration achieves the best performance, which we adopt in our research. As for our application, we note that it might not be the best choice to utilize all kinds of the features. From \autoref{fused features}, for example, we can see that the performance of incorporating three and four kinds of visual cues has an obvious decline. The reason is that the model falls into the overfitting problem when facing more features \cite{multimodal}. Specifically, a new TDCN branch is needed when adding another kind of visual cue, bringing in a significant increase of the parameter size for the whole model. As a result, the overfitting problem is inevitable given that the number of training subjects keeps on limited.
	
	\begin{table}[htbp]
		\caption{Results of different re-sampling strategies}
		\label{sample}
		\centering
		\begin{tabular}{ccccc}
			\hline
			\multicolumn{5}{c}{Validation Set}                                                                            \\ \hline
			\multicolumn{1}{c|}{Sample Method}                & Accuracy       & Recall         & Precision      & F1-score       \\ \hline
			\multicolumn{1}{c|}{\textbf{Head-first}} & \textbf{0.857}          & \textbf{0.917}          & \textbf{0.733} & \textbf{0.815}          \\
			\multicolumn{1}{c|}{Average}      & 0.629          & 0.583          & 0.467 & 0.519          \\ \hline
			\multicolumn{5}{c}{Test Set}                                                                               \\ \hline
			\multicolumn{1}{c|}{Sample Method}             & Accuracy       & Recall         & Precision      & F1-score       \\ \hline
			\multicolumn{1}{c|}{\textbf{Head-first}} & \textbf{0.660} & \textbf{0.643} & \textbf{0.450} & \textbf{0.530} \\
			\multicolumn{1}{c|}{Average}      & 0.617          & 0.571 & 0.400          & 0.471         \\ \hline

		\end{tabular}
	\end{table}
	
	We also investigate the effects of different re-sampling strategies, as shown in \autoref{sample}. In the data preprocessing stage, we empirically re-sample the sequence into a fixed length for all the training and test subjects. To represent a subject, Head-first means that we only use the a part of the sequence from the beginning. As for the Average strategy, the sequence of a subject is divided into pieces with the fixed length. The results based on this strategy are obtained through taking the average of the soft predicting scores of these pieces. From the table, we can see that the Head-first sampling strategy consistently performs better than the Average strategy in both sets. As the interviewing process is relatively long, many divided sub-sequences of one subject may not be depression-specific. In this context, the results based on the Average strategy can be inevitably lowered in a statistical sense.

    In the following, we conduct several ablation studies on the proposed model in terms of the backbone TDCN, the FWA module, and the max pooling operation. First, we replace TDCNs with TCNs in our model. The results in \autoref{tcntdcn} shows that TDCN significantly outperforms TCN in all the evaluation metrics. The results empirically validate the effectiveness of TDCN in aggregating global temporal information.	Besides, our TDCN has much lower FLOPs (4.5G vs 10.7G, the inputs are 2D Landmark sequences), showing its computational efficiency.

	\begin{table}[!htbp]
		\caption{Results of Different Backbones}
		\label{tcntdcn}
		\centering
		\begin{tabular}{ccccc}
			\hline
			\multicolumn{5}{c}{Validation Set}                                                                            \\ \hline
			\multicolumn{1}{c|}{Backbone}                & Accuracy       & Recall         & Precision      & F1-score       \\ \hline
			\multicolumn{1}{c|}{TCN} &  0.686        & 0.500         & 0.546   &     0.522      \\
			\multicolumn{1}{c|}{\textbf{TDCN}}      & \textbf{0.857}          & \textbf{0.917}          & \textbf{0.733} & \textbf{0.815}          \\ \hline
			\multicolumn{5}{c}{Test Set}                                                                               \\ \hline
			\multicolumn{1}{c|}{Backbone}             & Accuracy       & Recall         & Precision      & F1-score       \\ \hline
			\multicolumn{1}{c|}{TCN}     & 0.596     &  0.286    & 0.308 & 0.296 \\
			\multicolumn{1}{c|}{\textbf{TDCN}}      & \textbf{0.660}          & \textbf{0.643} & \textbf{0.450}          & \textbf{0.530}         \\ \hline
		\end{tabular}
	\end{table}

	To show the significance of the FWA module, we construct an incomplete model by not using FWA. In another word, the features learned from the TDCN branches are simply concatenated and directly sent into the fully connected layers. As is shown in \autoref{data}, compared with our model, the incomplete version suffers from the significant degeneration in terms of all four evaluation metrics, which verifies the usefulness of our fusion module assigning different weights to the feature channels.
	
    \begin{table}[!htbp]
		\caption{Results of Different Fusion Method}
		\label{data}
		\centering
		\begin{tabular}{ccccc}
			\hline
			\multicolumn{5}{c}{Validation Set}                                                                            \\ \hline
			\multicolumn{1}{c|}{Model}                & Accuracy       & Recall         & Precision      & F1-score       \\ \hline
			\multicolumn{1}{c|}{\textbf{With FWA}} & \textbf{0.857}          & \textbf{0.917}          & \textbf{0.733} & \textbf{0.815}          \\
			\multicolumn{1}{c|}{Without FWA}      & 0.771          & 0.583          & 0.700 & 0.636          \\ \hline
			\multicolumn{5}{c}{Test Set}                                                                               \\ \hline
			\multicolumn{1}{c|}{Model}             & Accuracy       & Recall         & Precision      & F1-score       \\ \hline
			\multicolumn{1}{c|}{\textbf{With FWA}} & \textbf{0.660} & \textbf{0.643} & \textbf{0.450} & \textbf{0.530} \\
			\multicolumn{1}{c|}{Without FWA}      & 0.575          & 0.143 & 0.200          & 0.167         \\ \hline
		\end{tabular}
	\end{table}
	
	In our model, the multiple max pooling layers gradually extract the maximum response from the long sequence. To validate this, we modify the model structure by replacing all the max pooling layers in TDCNs with the average pooling layers, and compare the performance based on these two pooling operations in \autoref{features}. We can see that the Max Pooling adopted in our model is on par with Average Pooling on the validation set, while it is slightly better than it counterpart on the test set. These results empirically show that it is more appropriate to choose the max pooling operation in the ADD task.
	
		
	\begin{table}[!htbp]
		\caption{Results of Different Pooling Layers}
		\label{features}
		\centering
		\begin{tabular}{ccccc}
			\hline
			\multicolumn{5}{c}{Validation Set}                                                                            \\ \hline
			\multicolumn{1}{c|}{Pooling Layer}                & Accuracy       & Recall         & Precision      & F1-score       \\ \hline
			\multicolumn{1}{c|}{Max Pooling} & 0.857          & 0.917          & 0.733 & 0.815          \\
			\multicolumn{1}{c|}{Average Pooling}      & 0.857          & 0.917          & 0.733 & 0.815          \\ \hline
			\multicolumn{5}{c}{Test Set}                                                                               \\ \hline
			\multicolumn{1}{c|}{Pooling Layer}             & Accuracy       & Recall         & Precision      & F1-score       \\ \hline
			\multicolumn{1}{c|}{\textbf{Max Pooling}} & \textbf{0.660} & \textbf{0.643} & \textbf{0.450} & \textbf{0.530} \\
			\multicolumn{1}{c|}{Average Pooling}      & 0.617          & \textbf{0.643} & 0.409          & 0.500          \\ \hline
		\end{tabular}
	\end{table}

	\section{Conclusion}
	In this paper, we conduct research on automatic depression detection (ADD) based on the visual cues recorded in the interviewing process. To fully explore the discriminating information from the sequence, we propose a novel ADD method that is mainly composed of Temporal Dilated Convolutional Networks (TDCN) and Feature-Wise Attention (FWA). The former tries to better extract the long-range temporal visual information from the sequences via dilated convolutions and max pooling at different scales. The later tries to fully utilize the different kinds of visual information reflecting facial details. We validate our method on the DAIC\_WOZ dataset. Extensive experimental results show the state-of-the-art performance over other related ADD methods, as well as the effectiveness of the elements in the proposed method.
	

\ifCLASSOPTIONcaptionsoff
  \newpage
\fi



%
    \bibliographystyle{IEEEtran}
    \bibliography{main}

\end{document}